# Population estimation using 3D city modelling and Carto2S datasets - A case study

Jai G Singla

Space Applications Centre, ISRO, Ahmedabad – 380015, India

*Abstract—* **With the launch of Carto2S series of satellites, high resolution images (0.6-1.0 meters) are acquired and available for use. High resolution Digital Elevation Model (DEM) with better accuracies can be generated using C2S multi-view and multi date datasets. DEMs are further used as an input to derive Digital terrain models (DTMs) and to extract accurate heights of the objects (building and tree) over the surface of the Earth. Extracted building heights are validated with ground control points and can be used for generation of city modelling and resource estimation like population estimation, health planning, water and transport resource estimations. In this study, an attempt is made to assess the population of a township using high-resolution Indian remote sensing satellite datasets. We used Carto 2S multi-view data and generated a precise DEM and DTM over a city area. Using DEM and DTM datasets, accurate heights of the buildings are extracted which are further validated with ground data. Accurate building heights and high resolution imagery are used for generating accurate virtual 3D city model and assessing the number of floor and carpet area of the houses/ flats/ apartments. Population estimation of the area is made using derived information of no of houses/ flats/ apartments from the satellite datasets. Further, information about number of hospital and schools around the residential area is extracted from open street maps (OSM). Population estimation using satellite data and derived information from OSM datasets can prove to be very good tool for local administrator and decision makers.**

INTRODUCTION

Across the globe, there are many satellites in orbit such as Worldview series, Pleiades and Ikonos, those are acquiring very high resolution datasets over the Earth. India has also launched a series of Carto2S satellites, which captures 60 cm to 1-meter data over different parts of the World. Data is available in panchromatic as well as in multispectral range. Satellite has capabilities to acquire multi-view and multi date datasets. These multi-view datasets from different view angles are handy for generation of accurate DEM using the technique of satellite photogrammetry. By taking inputs from such high resolution datasets, initial estimation for various resources can be made. In this study, we are estimating Population of small area using the C2S datasets.

Many researchers have attempted to estimate population using satellite data by employing the techniques of areal interpolation or statistical modeling based methods. Researchers in early 2000s tried to estimate population on 2D image datasets with a spatial resolution of 5-10 meters. Population estimation using 2D datasets performs better in homogenous areas but it gives significant errors in case of urban areas where buildings have variation in the number of floors. Therefore, over a time with the launch of high resolution satellite datasets, researchers across the world are trying to estimate population using combination of accurate DSM, high resolution image datasets and 3D city models.

In year 2005, Sheng Wu and his team provided a descriptive review of population estimation methods (Shen Wu, 2005) . They provided detailed information about areal interpolation methods which uses census population data as an input and statistical methods which infers the relationship between population and other variables for estimating population. MD Islam et al. at Fabspace lab tried to map human settlement to estimate population using sentinel-2 data (MD Islam, 2017). They mainly classified the sentinel-2 image into classes of build-up area, vegetation, water and cloud. Using the information of build-up area, polygon layer and population density factor, they tried to estimate population of the Lusaka and Uganda. They had the limitation of measuring the number of floors of the buildings due to 10m resolution of data.

In year 2016, Filip Biljecki et al. worked on population estimation using a 3D city model and estimated the population of Netherlands (Filip Biljecki, 2016). They concluded that 3D city models have clear advantage over 2D approach. Team from Singapore in year 2017 examined Google earth image of spatial resolution 2.15 meters to estimate population of west coast in Singapore with an accuracy of 92.5 % (K. Cao, 2017). In year 2019, Roger Hilson et. al. used Landsat 30m imagery to estimate the size of urban population in building structures using Bayesian methods over Bo, Sierra, Leone, West Africa (R. Hilson, 2019). They claimed to estimate total population of 20 neighborhoods with an error of less than 1 percent.

Since, most of the research worked on well planned and clean structures using foreign datasets. In this study, we have taken a residential society of Ahmedabad city of India. For the very first time, an initial attempt is made to estimate population over Ahmedabad city using Indian remote sensing satellite datasets of high resolution images and multi-view data. We have generated DEM, DTM and then 3D city model over the study area and using volume based approach, estimated the population of a small society. Further, we validated our results with available ground truth. Further, using open street map datasets, number of hospital and schools in the study area is also mapped.

STUDY SITE AND DATASETS

a. High resolution Imagery: Multi spectral high resolution image (60cm) from Carto 2S mission.
b. Multi-view data: Pair of multi-view images with 60cm resolution from C2S with following details:
Product ID = 16923837111

Date of Pass = Oct 2016

Orbit No. = 1565

Product ID = 16923837131

Date of Pass = Nov 2016

Orbit No. = 2066

c. Open street map files: Open street maps data over study area (~5m spatial resolution) of year 2018-19 was downloaded from GeoFabrik website (Geofabrik team). Further, information of number of schools and hospital around the residential society was extracted from open street map datasets using rule based filtering.

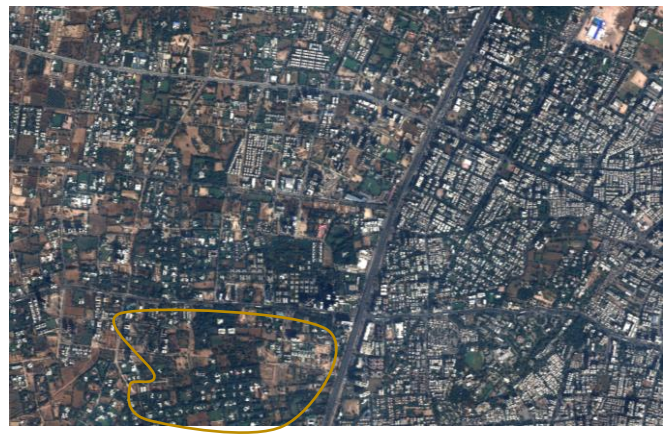

**Figure -1 City of Ahmedabad (1-meter multispectral image 23.0225-degree North, 72.5-degree East)**

METHODOLOGY

In order to make accurate estimation of population in the study area, we have generated digital elevation model using high resolution satellite images using the approach of satellite photogrammetry. Further, processed DEM to obtain digital terrain model (DTM) by removing the heights of the objects over Earth surface. Heights of the building structures were obtained by subtracting the DEM and DTM height over the building and are recorded for 3D city modelling and floor height assessment. Detailed steps are listed below:

a) **Generation of DEM using multi-view imagery over the study area**

Accurate digital elevation models play a very significant role in estimation of accurate population of the area. we have taken multi-view multi-date (Oct 2016 and Nov 2016) datasets from Carto2S satellite and generated DEM over study area as per approach mentioned in our earlier publication (Singla, 2021, To be published).

b) **Extraction of heights from building footprints**

For volume based assessment of the building area, it is important to extract heights of the individual building units. After generating a fine DEM from the first step, bare earth surface (DTM) is derived using QGIS slope filtering toolbox. Using DEM and DTM of the study area, height of the individual building is extracted and an accurate city model is generated for that residential society. Details about this step is listed in our earlier publications (Singla, 2021, To be published).

**c) Population estimation using available satellite data**

The most crucial part of the paper is derivation of population estimation using image data and extracted height values for each building. We have used following steps to estimate population of the area:

**Population of the Area= No of units' x No of floors x occupancy rate**

**No of floors= height of the building/ height of one floor**

**No of units = total carpet area of the building / carpet area of single unit**

Here, height of the buildings is derived by processing the DEM and DTM information. Entire building footprint is measured from high resolution image and in the same way, single apartment built up area is also calculated. we need to have local information about occupancy rate to make more accurate estimates.

**d) Open street map data and extraction of relevant information**

Geofabrik provides open sources datasets for full globe coverage. These datasets can be categorized under buildings, roads, places, amenities, parks and water ways and many more categories. We have extracted the important information of number of buildings, schools and hospital from OSM and used it to generate and visualize city models using approach mentioned in our earlier publication (Singla, 2020). Further, statistics of hospitals and schools were extracted to establish resources linkage with number of residents in the studied society.

RESULTS

In this section, we are describing the results obtained using satellite data. Methodology mentioned in this section 3 is applied on the study area for each of the building unit. Following table depicts the extracted height of the building and carpet area of the building units based on building type.

**Table-1 Estimated height of the Type-1 buildings and no of units based on satellite data**

| Sr. no | Extracted Height of the building | Calculated No of floors | Build up area of entire building (Length x breadth) | Built up area of one unit (Length x breadth) | No of units per tower | Remarks |
|---|---|---|---|---|---|---|
| 1. | 19.8 m | >6 (7) | 30x30.5 meters | 15x12.5 meter | 4 | Type1 |
| 2. | 19.8 m | > 6 (7) | 30x29.8 meters | 15x12 meter | 4 | |
| 3. | 20m | > 6 (7) | 30x27 meters | 15x11.8 meter | 4 | |
| 4. | 4.5m | >1(2) | 28x28 meter | 13x11.2 meter | 4 | |
| 5. | 4.5m | >1 (2) | 28x28 meter | 13x11.2 meter | 4 | |
| 6. | 4.5m | >1 (2) | 28x28 meter | 13x11.2 meter | 4 | |
| 7. | 19.3 m | >6 (7) | 28.5x 29 m | 13x 11.2 m | 4 | |

| | | | | | 136 | |
|---|---|---|---|---|---|---|
| Total no of units | | | | | | |

**Table-2 Estimated height of the Type-2 buildings and no of units based on satellite data**

| Sr. no | Extracted Height of the building | Calculated No of floors | Build up area of entire building (Length x breadth) | Built up area of one unit (Length x breadth) | No of units per tower | Remarks |
|---|---|---|---|---|---|---|
| 1. | 19.5 m | >6 (7) | 19x27.2 meters | 9.5x9.5 meter | 4 | Type2 |
| 2. | 19.5 m | > 6 (7) | 19x27.2 meters | 9.5x9.5 meter | 4 | |
| 3. | 19.8 m | > 6 (7) | 19x27.2 meters | 9.5x9.5 meter | 4 | |
| 4. | 5.6m | >1 (2) | 20.5x31 meter | 8.5x10 meter | 4 | |
| 5. | 5.0m | >1 (2) | 20.5x31 meter | 8.5x10 meter | 4 | |
| 6. | 4.85m | >1 (2) | 20.5x31 meter | 8.5x10 meter | 4 | |
| 7. | 4.8 m | >1 (2) | 20.5x31 meter | 8.5x10 meter | 4 | |

| Sr. no | Extracted Height of the building | Calculated No of floors | Build up area of entire building | Built up area of one unit | No of units per tower | Remarks |
|---|---|---|---|---|---|---|
| 8. | 4.2 m | >1 (2) | 20.5x31 meter | 8.5x10 meter | 4 | |
| 9. | 19.5m | >6 (7) | 18.5x27.7 meter | 9x9.9 meter | 4 | |
| 10. | 19.2m | >6 (7) | 18.5x27.7 meter | 9x9.9 meter | 4 | |
| 11. | 19.7 m | >6 (7) | 18.5x27.7 meter | 9x9.9 meter | 4 | |
| Total no of units | | | | | 208 | |

**Table-3 Estimated height of the Type3 building and no of units based on satellite data**

| Sr. no | Extracted Height of the building | Calculated No of floors | Build up area of entire building | Built up area of one unit | No of units per tower | Remarks |
|---|---|---|---|---|---|---|
| 1. | 11 m | >3 (4) | 17x17.5 | 8x6.5 | 4 | Type 3 |
| 2. | 9 m | >2 (3) | 17x17.5 | 8x6.5 | 4 | |
| 3. | 8 m | >2 (3) | 17x17.5 | 8x6.5 | 4 | |
| 4. | 8 m | >2 (3) | 17x17.5 | 8x6.5 | 4 | |
| 5. | 8 m | >2 (3) | 17x17.5 | 8x6.5 | 4 | |
| 6. | 10 m | >3 (3) | 17x17.5 | 8x6.5 | 4 | |
| 7. | 8 m | >2 (3) | 17x17.5 | 8x6.5 | 4 | |
| 8. | 8 m | >2 (3) | 17x17.5 | 8x6.5 | 4 | |
| *9.* | 10 m | >3 (4) | 17x17.5 | 8x6.5 | 4 | |
| *10.* | 12 m | >3 (4) | 17x17.5 | 8x6.5 | 4 | |
| *11.* | 10 m | >3 (4) | 17x17.5 | 8x6.5 | 4 | |
| *12.* | 12 m | >3 (4) | 20x17 | 8.5x6 | 4 | |
| *13.* | 12 m | >3 (4) | 20x17 | 8.5x6 | 4 | |
| *14.* | 11 m | >3 (4) | 20x17 | 8.5x6 | 4 | |
| *15.* | 6 m | >1 (2) | 20x17 | 8.5x6 | 4 | |
| *16.* | 10 m | >3 (4) | 20x17 | 8.5x6 | 4 | |
| *17.* | 12 m | >3 (4) | 20x17 | 8.5x6 | 4 | |
| **Total no of units** | | | | | **240** | |

**Table-4 Estimated height of the Type4 building and no of units based on satellite data**

| Sr. no | Extracted Height of the building | Calculated No of floors | Build up area of entire building | Built up area of one unit | No of units per tower | Remarks |
|---|---|---|---|---|---|---|
| *1.* | 10 m | >3 (4) | 15x16.5 meters | 7.2x6 meter | 4 | Type 4 |
| *2.* | 11 m | > 3 (4) | 15x16.5 meters | 7.2x6 meter | 4 | |
| **Total no of units** | | | | | **32** | |

**Table-5 Estimated height of the Type5 building and no of units based on satellite data**

| Sr. no | Extracted Height of the building | Calculated No of floors | Build up area of entire building | Built up area of one unit | No of units per tower | Remarks |
|---|---|---|---|---|---|---|
| 1. | 10 m | >3 (4) | 14x16 meters | 6.5x5.5 meter | 4 | Type-5 |
| 2. | 10 m | > 3 (4) | 14x16 meters | 6.5x5.5 meter | 4 | |
| Total no of units | | | | | 32 | |

## VALIDATIONS

In this section, we are explaining the validation part of our study. we have taken the actual ground data for this residential society for all the five types of apartments and compared the results of satellite based estimation to ground data as per the following table. Estimated building heights are very close to actual building heights of the society. Based on the building heights and carpet area, total no of flats as per type of the flat is estimated as per Table-6. As per the difference of percentage metric, estimate accuracies are <3 percent. Further to population estimates, resources such as no of hospitals and no of schools are inferred from OSM data and conclusion of availability of resources can be drawn using satellite datasets.

**Table-6 Validation of estimated number of unit's Vs no of units in actual ground data**

| Sr. No | Type of apartment | Estimated no of units | Ground data | Diff in percentage of no of units |
|---|---|---|---|---|
| 1. | Type 1 | 136 | 133 | 2.25 |
| 2. | Type 2 | 208 | 211 | 1.44 |

| 3. | Type 3 | 240 | 241 | 0.41 |
| 4. | Type 4 | 32 | 43 | 34.35 |
| 5. | Type 5 | 32 | 31 | 3.1 |
| 6. | **Total** | **648** | **659** | **1.66** |

So, as per the comparisons of estimated datasets to ground datasets, it is inferred that we have achieved very good accuracy of detection of no of units over this residential township. We have an error percent of 1.66 which is very low. Hence, usage of high resolution images and DEM data is very useful in applications of estimation of numbers of apartments in an area. We have also compared the estimated built-up area of units with the ground data and obtained more than 98% accuracy in the estimations.

In order to draw conclusions of number of residents in the society, one needs information about occupancy rate and population density like local knowledge. But, in this case, without using any local information and by using the knowledge of apartment built up area extracted using satellite data, we have predicted population of this residential society as per the following Table-6. We have used the knowledge from Table-1 to 5 for this estimate.

**Table-6 Estimation of total no of persons in the society**

| Unit type | Built up area of 1 unit | Population assessment | Total population estimate |
|---|---|---|---|
| Type 1 | 146-187 sq meters | 6 persons / unit | 816 |
| Type 2 | 85-90 sq meters | 4 persons/ unit | 816 |
| Type 3 | 50-52 sq meters | 3-4 persons / unit | 840 |
| Type 4 | 43.1 sq meters | 3 persons / unit | 96 |
| Type 5 | 35.75 sq meters | 2 person/ unit | 64 |
| **Total** | | | **2632** |

As we have deduced the estimation of no of persons in a residential society using satellite data and similarly this method can be applied to calculate population estimate for relatively larger area, further for a town level and then for a very big city. Further, after every 2 years' population estimate can be generated to estimate the increase in population of an area and requirements of public resources for local population. For more accurate estimates of the population, local knowledge about occupancy rate and population density is very important. In our study , we have used occupancy rate of 1 on the scale of (0-1) as per local information.

Once, population of the area is calculated, important resources like no of hospitals, no of schools and no of police station etc. can be mapped to the area of interest. In this study, we have mapped the number of hospitals and number of schools around the studied residential society using the datasets from publically available OSM files provided by geofabrik. In the following figures, it can be inferred that within 12 km area from the residential society, there are at least 25-30 hospitals and more than 6 schools for kids. So, by this kind of analysis we can get the complete information about the number of residents and resources shared among them using satellite datasets.

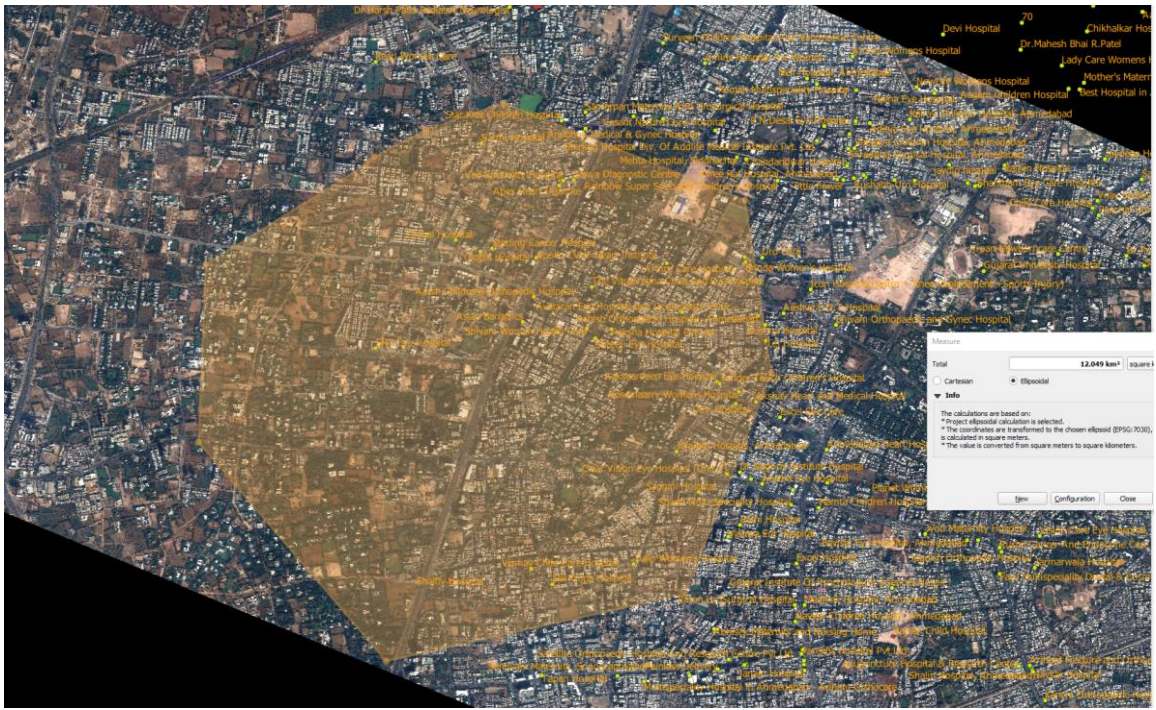

Figure-1: Number of Hospitals falls within12 km Sq meter Area of residential society

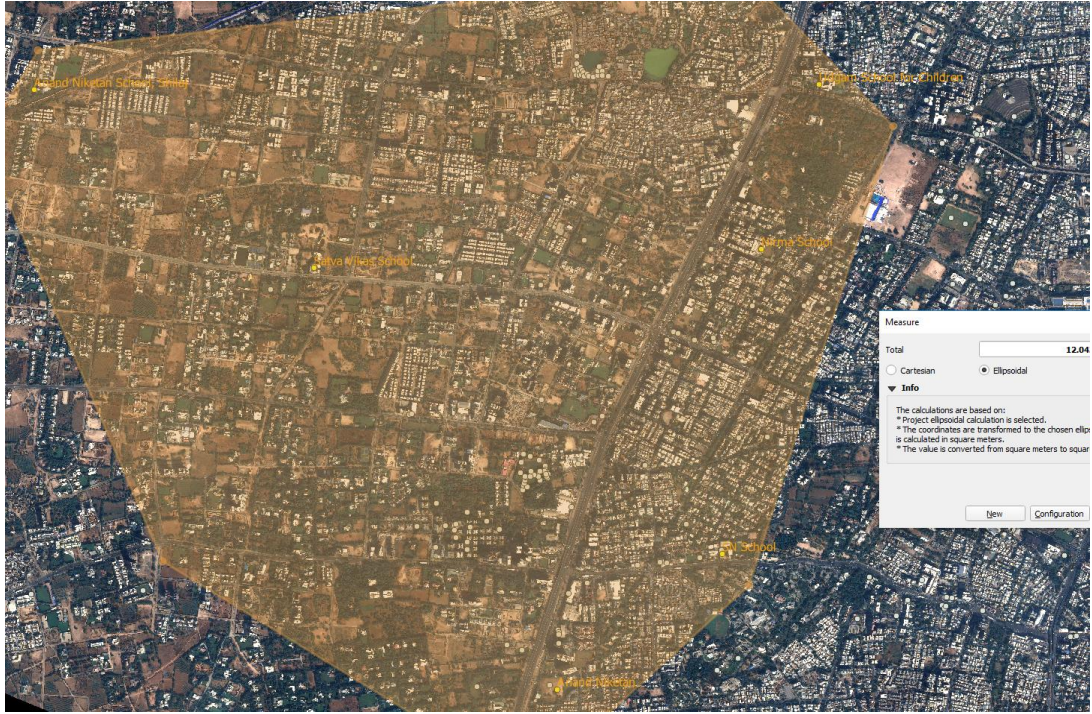

Figure-2: Number of Schools falls within 12 km Sq meter Area of residential society

LIMITATIONS AND CHALLENGES

Population estimation task using satellite data involves a lot of parameters as it is very dynamic in nature. In the above study, we have estimated no of buildings and height of the buildings in a residential society using high resolution satellite data with very good accuracies (98-99%). Heights of the buildings derived using satellite data matched with ground truth with an accuracy of more than 95% in our case. So, in this study, we are able to capture floor level changes. In case, extracted heights are not so accurate or it has error of more than 3m, there will be inaccuracies in calculation of number of flats and it will carry errors in population estimation too.

There is scope of generating of more accurate (<1m) DEM using high resolution satellite data and fine resolution image which will help to get more accurate estimates of the no of units in the societies and scope of error will further diminish. Because of dynamic nature, Population estimation further requires local knowledge of occupancy ratio and rough estimates of density rate of the area to predict the very accurate number of people residing in the society.

In this study, we have done the population estimation for a limited area because of lack of resources and limited validation datasets. Population estimation for a bigger city is feasible using satellite data in case enough resources are made available. Another challenge is to perform population estimation automatically using satellite data without using any manual methods. People have attempted automatic population estimation using conventional methods where they rely mainly on previous census datasets or other information. Known population of bigger area is narrowed down for a smaller area based on density ratio and zonal area mapping where a lot of heuristic and input parameters are required. Currently, there is no automatic way to estimate population based on volume based approach.

## CONCLUSIONS

In this study, we have taken very high resolution datasets from C2S satellite and generated very accurate DEM and DTM of the study area. Using the satellite information, we have extracted the heights of the individual buildings over the study area and also calculated the built-up area for one apartment as well as for the entire building footprint. We have used extracted knowledge to estimate the number of units in a residential society and we got very encouraging results of better than 98% accuracy. Further, based on the built-up areas of the apartments and without using any local information, we have estimated the no of persons living in the particular society with very good accuracy. The accuracy of population estimate of studied society is also very encouraging. Further, using open street map datasets, we have mapped the important resources of hospitals and schools with the residential area. Mapping of resources based on the total population estimates of the residential areas will be very useful tool for administrators and policy makers to use. We draw the conclusion that that population estimates with good accuracies are possible using satellite data and it will be useful to map various public resources for local administrators and policy makers for effective decision making.